\pdfoutput=1

\PassOptionsToPackage{table,xcdraw}{xcolor} 

\documentclass[11pt]{article}

\usepackage[final]{acl}

\usepackage{times}
\usepackage{latexsym}

\usepackage[T1]{fontenc}

\usepackage[utf8]{inputenc}

\usepackage{microtype}

\usepackage{inconsolata}

\usepackage{graphicx}

\usepackage{amsmath}
\usepackage{multirow}
\usepackage{xcolor}

\usepackage{booktabs}
\usepackage{tabularray}

\usepackage{algorithm}
\usepackage{algpseudocode}

\newcommand\blfootnote[1]{%
  \begingroup
  \renewcommand\thefootnote{}\footnote{#1}%
  \addtocounter{footnote}{-1}%
  \endgroup
}

\title{Exploring \textit{Coding Spot}:\\Understanding Parametric Contributions to LLM Coding Performance}

\author{Dongjun Kim$^{\star}$, Minhyuk Kim$^{\star}$, YongChan Chun, Chanjun Park$^{\dagger}$, Heuiseok Lim$^{\dagger}$ \\
\\
  Korea University \\
  \texttt{\{junkim100, mhkim0929, cyc9805, bcj1210, limhseok\}@korea.ac.kr}}

\begin{document}
\maketitle
\blfootnote{$^\star$ Equal contribution}
\blfootnote{$^\dagger$ Corresponding Author}
\begin{abstract}
Large Language Models (LLMs) have demonstrated notable proficiency in both code generation and comprehension across multiple programming languages. However, the mechanisms underlying this proficiency remain underexplored, particularly with respect to whether distinct programming languages are processed independently or within a shared parametric region. Drawing an analogy to the specialized regions of the brain responsible for distinct cognitive functions, we introduce the concept of \textit{Coding Spot}, a specialized parametric region within LLMs that facilitates coding capabilities. Our findings identify this \textit{Coding Spot} and show that targeted modifications to this subset significantly affect performance on coding tasks, while largely preserving non-coding functionalities. This compartmentalization mirrors the functional specialization observed in cognitive neuroscience, where specific brain regions are dedicated to distinct tasks, suggesting that LLMs may similarly employ specialized parameter regions for different knowledge domains.
\end{abstract}

\section{Introduction}
Large Language Models (LLMs) have initiated a significant transformation in computational code processing, showcasing advanced capabilities in tasks such as code generation and comprehension across a wide range of programming languages \cite{chen2021evaluating, austin2021program, li2022competition}. Models such as Llama 3 \cite{dubey2024llama}, GPT-4o \cite{achiam2023gpt}, and Claude 3.5 Sonnet \cite{Anthropic_2024} have achieved considerable success, establishing themselves as essential tools for automating programming tasks and enhancing developer productivity.

Despite these successes, a fundamental question remains: how do these models internally represent and organize the coding knowledge necessary for such tasks? More specifically, it is unclear whether the knowledge required for programming tasks is uniformly distributed across the model’s parameters or if certain parameter subsets exhibit specialization for coding-related functionalities. This question is reminiscent of findings in cognitive science, where specific brain regions, such as Broca’s and Wernicke’s areas, are specialized for language processing \cite{broca1861remarks, wernicke1874aphasische}. Inspired by this analogy, we hypothesize that LLMs may similarly exhibit task-specific parametric regions, particularly those dedicated to coding tasks.

In this study, we introduce the concept of the \textit{Coding Spot}, a theoretical construct representing a subset of parameters within LLMs that are particularly critical for code-related capabilities. Analogous to domain-specific regions in the brain, the \textit{Coding Spot} embodies a specialized parametric region that is crucial for the model’s proficiency in coding tasks. By identifying and analyzing these critical regions, we aim to provide new insights into the internal parametric architecture of LLMs and their ability to manage coding and general tasks.

Our primary objective is to conduct a rigorous examination of the parametric structure of LLMs, with a focus on uncovering the role of the \textit{Coding Spot}. We evaluate the consequences of modifying this subset of parameters on both coding and non-coding tasks, shedding light on the compartmentalized nature of the LLMs' internal architecture. This investigation offers a deeper understanding of how LLMs handle domain-specific knowledge and draws compelling parallels to the cognitive specialization observed in the human brain.

\begin{figure}[t]
\resizebox{1.1\columnwidth}{!}{
  \includegraphics[width=\columnwidth]{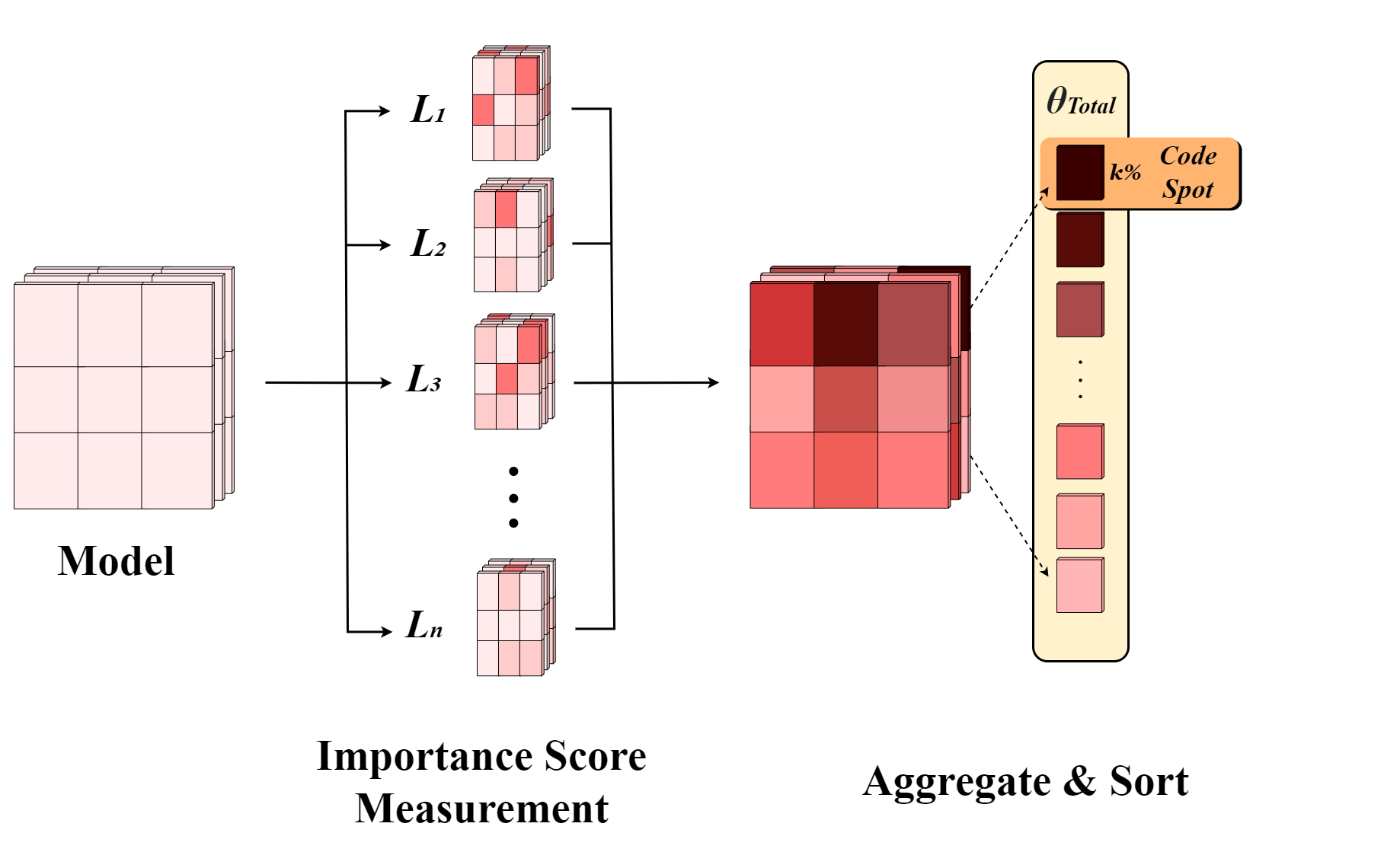}
}
  \caption{Overview of the framework for extracting and analyzing Coding Spot within LLMs. The process begins with the model undergoing importance scoring independently for n programming languages. The Importance Scores extracted from each language are aggregated for each parameter and then sorted in descending order. The parameters within the top k\% of Importance Scores are defined as the Coding Spot.}
  \label{fig:code-flowchart}
\end{figure}

\section{Coding Spot}
Our methodology aims to identify and analyze the \textit{Coding Spot}, a specialized subset of parameters crucial for coding proficiency in LLMs. 

\paragraph{Methodological Framework}
The core of our methodology is a systematic algorithm designed to identify critical subsets of parameters—termed the \textit{Coding Spot}—from a large pool of LLM parameters. By employing parameter importance scoring, our framework efficiently isolates the parameters most relevant to coding tasks, drawing analogies to the specialization observed in the human brain, where distinct regions are responsible for different cognitive functions.

\paragraph{Parameter Importance Scoring}
\label{sec:parameter_importance}

To isolate the \textit{Coding Spot}, we begin by fine-tuning LLMs on datasets containing code from individual programming languages. The purpose of this fine-tuning is not to build a new task-specific model but to extract accurate parameter gradients via backpropagation. These gradients allow us to construct language-specific parameter subsets that are vital for coding tasks.

Given a dataset \( D_l \) corresponding to a programming language \( l \) and a set of parameters \( \theta = [\theta_1, \theta_2, \ldots, \theta_d] \), our goal is to estimate how changes in the loss function \( L(D_l, \theta) \) relate to each parameter. Using the first-order Taylor expansion, the model's loss in response to a specific parameter \( \theta_j \) can be approximated as:

\begin{equation}
L(D_l, \theta) \approx L(D_l, \theta|_{\theta_j = 0}) + \frac{\partial L(D_l, \theta)}{\partial \theta_j} \cdot \theta_j
\end{equation}

This equation highlights how the loss is affected by parameter \( \theta_j \), informed by its gradient during fine-tuning. The parameter importance score \( I^l_j(\theta) \), which quantifies each parameter's contribution to coding tasks, is computed as:

\begin{equation}
I^l_j(\theta) \approx \left| \frac{\partial L(D_l, \theta)}{\partial \theta_j} \right| \cdot \left| \theta_j \right|
\end{equation}

This score provides a direct measure of each parameter's relevance in the context of a specific programming language \( l \), revealing the most critical parameters for coding tasks. The role of the fine-tuned model here is solely to facilitate precise gradient extraction, not to be used in subsequent analyses.

\paragraph{Aggregating Parameter Importance Across Languages}
Once we calculate the importance scores \( I^l_j(\theta) \) for each parameter within individual languages, the next step is to aggregate these scores across multiple languages. For each parameter \( \theta_j \), we compute a total importance score \( I^{\text{total}}_j(\theta) \) by summing the importance scores across all languages in the set \( L \):

\begin{equation}
I^{\text{total}}_j(\theta) = \sum_{l \in L} I^l_j(\theta)
\end{equation}

This aggregation captures the global importance of each parameter, allowing us to identify parameters that consistently influence coding tasks across diverse languages. By sorting the parameters in descending order based on their total importance scores, we isolate a subset of parameters—referred to as the \textit{Coding Spot}—that are crucial for coding proficiency.

\paragraph{Defining the \textit{Coding Spot}}
The \textit{Coding Spot} is identified by selecting the top \( k\% \) parameters from the sorted list. These parameters, which consistently demonstrate high importance across multiple languages, form a concentrated subset responsible for coding tasks. The value of \( k \) is empirically determined to ensure that we capture the most critical parameters while avoiding redundancy.

\begin{table*}[ht]
\centering
\resizebox{0.8\linewidth}{!}{

    \begin{tabular}{lccccccc}
        \toprule
                       & \multicolumn{6}{c}{\textbf{General}}                                                   & \textbf{Code}   \\
                       & \textbf{GSM8K} & \textbf{HellaSwag} & \textbf{MMLU} & \textbf{TruthfulQA} & \textbf{WinoGrande} & \textbf{Avg. GTC (\%)} & \textbf{HumanEval} \\
        \toprule
        \multicolumn{8}{c}{\textbf{\textit{CodeLlama 7B Instruct}}}                                                                                           \\
        \midrule
        \textbf{Original}       & 18.12  & 48.32  & 39.54  & 39.21  & 64.56  & 41.95  & 87.2  \\
        \textbf{\includegraphics[height=10.pt]{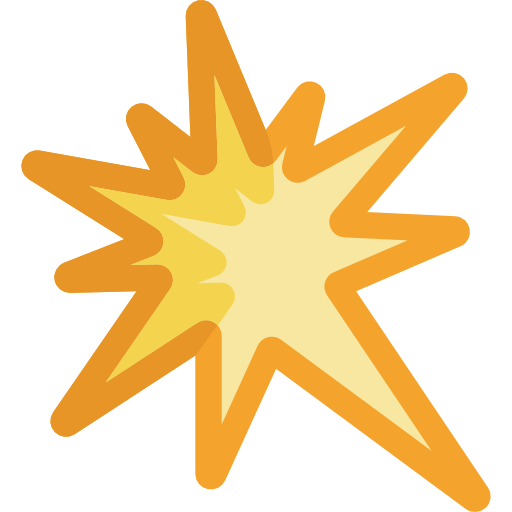}  0.0025\%} & 1.90   & 36.01  & 24.63  & 39.32  & 53.51  & \cellcolor[HTML]{fff6f6}31.07 (-25.93\%) & \cellcolor[HTML]{ffadaa}22.56 (-74.13\%) \\
        \textbf{\includegraphics[height=10.pt]{icons/boom.png}  0.01\%}  & 2.27   & 35.08  & 23.99  & 36.91  & 52.41  & \cellcolor[HTML]{fff6f6}30.13 (-28.17\%) & \cellcolor[HTML]{ff8e8a}16.46 (-81.12\%) \\
        \textbf{\includegraphics[height=10.pt]{icons/boom.png}  0.09\%}  & 0.23   & 27.23  & 22.94  & 49.72  & 51.38  & \cellcolor[HTML]{fff6f6}30.30 (-27.77\%) & \cellcolor[HTML]{ff6e69}1.83 (-97.90\%) \\
        \textbf{\includegraphics[height=10.pt]{icons/boom.png}  0.25\%}  & 0.00   & 25.85  & 22.93  & 51.52  & 50.51  & \cellcolor[HTML]{fff6f6}30.16 (-28.10\%) & \cellcolor[HTML]{ff4848}0.00 (-100.00\%) \\
        \toprule
        \multicolumn{8}{c}{\textbf{\textit{Llama 3.1 8B Instruct}}}  \\
        \midrule
        \textbf{Original}       & 76.72  & 59.10  & 67.96  & 54.08  & 73.64  & 66.30  & 97.56  \\
        \textbf{\includegraphics[height=10.pt]{icons/boom.png}  0.0025\%} & 2.35   & 44.47  & 42.32  & 43.61  & 60.14  & \cellcolor[HTML]{ffe7e7}38.58 (-41.81\%) & \cellcolor[HTML]{ffadaa}20.12 (-79.38\%) \\
        \textbf{\includegraphics[height=10.pt]{icons/boom.png}  0.01\%}  & 2.65   & 38.83  & 29.47  & 41.55  & 58.41  & \cellcolor[HTML]{ffe7e7}34.18 (-48.44\%) & \cellcolor[HTML]{ff6e69}9.76 (-90.00\%) \\
        \textbf{\includegraphics[height=10.pt]{icons/boom.png}  0.09\%}  & 0.61   & 26.82  & 23.57  & 49.58  & 49.17  & \cellcolor[HTML]{ffcac9}29.95 (-54.83\%) & \cellcolor[HTML]{ff4848}0.00 (-100.00\%) \\
        \textbf{\includegraphics[height=10.pt]{icons/boom.png}  0.25\%}  & 0.15   & 26.36  & 22.95  & 51.03  & 48.46  & \cellcolor[HTML]{ffcac9}29.79 (-55.07\%) & \cellcolor[HTML]{ff4848}0.00 (-100.00\%) \\
        \toprule
        \multicolumn{8}{c}{\textbf{\textit{Llama 3.2 3B Instruct}}}                                                                                          \\
        \midrule
        \textbf{Original}       & 64.67  & 52.22  & 60.42  & 49.77  & 67.56  & 58.93  & 94.51  \\
        \textbf{\includegraphics[height=10.pt]{icons/boom.png}  0.0025\%} & 2.35   & 41.39  & 38.71  & 45.07  & 55.96  & \cellcolor[HTML]{fff6f6}36.70 (-37.73\%) & \cellcolor[HTML]{ff8e8a}15.24 (-83.87\%) \\
        \textbf{\includegraphics[height=10.pt]{icons/boom.png}  0.01\%}  & 2.20   & 36.31  & 30.30  & 45.66  & 53.75  & \cellcolor[HTML]{ffe7e7}33.64 (-42.91\%) & \cellcolor[HTML]{ff6e69}6.71 (-92.90\%) \\
        \textbf{\includegraphics[height=10.pt]{icons/boom.png}  0.09\%}  & 1.29   & 26.66  & 23.36  & 50.53  & 49.49  & \cellcolor[HTML]{ffe7e7}30.27 (-48.64\%) & \cellcolor[HTML]{ff4848}0.00 (-100.00\%) \\
        \textbf{\includegraphics[height=10.pt]{icons/boom.png}  0.25\%}  & 1.21   & 26.21  & 22.94  & 49.54  & 49.96  & \cellcolor[HTML]{ffe7e7}29.97 (-49.14\%) & \cellcolor[HTML]{ff4848}0.00 (-100.00\%) \\
        \bottomrule
    \end{tabular}

}
\caption{Performance comparison of three LLMs on general and coding tasks after deactivating a percentage of the model identified as \textit{Coding Spot}. GTC represents General Task Change, and \includegraphics[height=8.pt]{icons/boom.png} marks models with deactivated parameters.}
\label{tab:results1}
\end{table*}

\section{Experiments}
The objective of our experimental evaluation is to quantify the role and impact of the \textit{Coding Spot} in LLMs, particularly its influence on both task-specific (e.g., coding) and general tasks (e.g., mathematical or commonsense reasoning). We systematically deactivate the identified \textit{Coding Spot} parameters to examine their specialization and robustness, thus providing insights into the broader implications of parameter specialization in LLMs.

\subsection{Experimental Setup}

The experimental setup was carefully designed to ensure comprehensive evaluation of both specialized and general model performance. Detailed information about the datasets, models, and evaluation benchmarks can be found in Appendix \ref{appen:experimental_setup}.

\paragraph{Datasets} Our study employed carefully curated training datasets and evaluation benchmarks to rigorously assess both specialized and general performance aspects of the models. For the fine-tuning phase, models were trained on the \texttt{nampdn-ai/tiny-codes} \cite{nam_pham_2023} dataset, meticulously filtered to include only pure code, thus excluding any content related to instruction following or English language capabilities. This focus was crucial to exclude instruction-following and English abilities, ensuring that models could extract and focus only on coding skills.

For code-related task evaluation, we utilized the HumanEval benchmark \cite{chen2021evaluating}, which assesses the models' capabilities in generating correct code solutions across a range of programming challenges. To evaluate general task proficiency, we employed a diverse set of benchmarks including GSM8K \cite{cobbe2021training} (assessing mathematical reasoning), HellaSwag \cite{zellers2019hellaswag} (for commonsense reasoning), and MMLU \cite{hendrycks2020measuring} (for multi-task evaluation), TruthfulQA \cite{lin2021truthfulqa} (to assess truthfulness of responses), and WinoGrande \cite{sakaguchi2021winogrande} (for coreference reasoning).

\paragraph{Models} We evaluated three state-of-the-art LLMs of varying sizes: CodeLlama 7B Instruct \cite{roziere2023code}, Llama 3.1 8B Instruct, and Llama 3.2 3B Instruct \cite{dubey2024llama}. These models were chosen to assess how different architectures and scales influence the identification and impact of the \textit{Coding Spot}. During fine-tuning, Python was excluded to test whether the \textit{Coding Spot} extends its functionality beyond language-specific constraints, thereby assessing its generalization across different coding environments.

\paragraph{Evaluation Metrics} The primary metric for code-related tasks was the HumanEval score, while general tasks were evaluated using accuracy metrics on GSM8K, HellaSwag, and MMLU. The key experimental procedure involved systematically nullifying varying percentages of the \textit{Coding Spot} parameters and evaluating their impact on both types of tasks. By excluding Python during fine-tuning, we tested the hypothesis that the \textit{Coding Spot} reflects a broader coding proficiency, generalizable across different programming languages.

\begin{table}[t]
\centering
\resizebox{\linewidth}{!}{

\begin{tabular}{lccc}
\toprule
               & \textbf{Avg. GTC (\%)} & \textbf{Code Change (\%)} & \(\boldsymbol{M_s}\)
 \\
\midrule
\multicolumn{4}{c}{\textbf{\textit{CodeLlama 7B Instruct}}}                                                        \\
\midrule
\textbf{Original}       & 41.95                     & 87.20               & -                            \\
\textbf{\includegraphics[height=10.pt]{icons/boom.png}  0.0025\%} & -25.93\%                  & -74.13\%            & 5.44                         \\
\textbf{\includegraphics[height=10.pt]{icons/boom.png}  0.01\%} & -28.17\%                  & -81.12\%            & 5.52                         \\
\textbf{\includegraphics[height=10.pt]{icons/boom.png}  0.09\%} & -27.77\%                  & -97.90\%            & 6.75                         \\
\textbf{\includegraphics[height=10.pt]{icons/boom.png}  0.25\%} & -28.10\%                  & -100.00\%           & 6.82                         \\
\toprule
\multicolumn{4}{c}{\textbf{\textit{Llama 3.1 8B Instruct}}}                                                        \\
\midrule
\textbf{Original}       & 66.30                     & 97.56               & -                            \\
\textbf{\includegraphics[height=10.pt]{icons/boom.png}  0.0025\%} & -41.81\%                  & -79.38\%            & 2.70                         \\
\textbf{\includegraphics[height=10.pt]{icons/boom.png}  0.01\%} & -48.44\%                  & -90.00\%            & 2.65                         \\
\textbf{\includegraphics[height=10.pt]{icons/boom.png}  0.09\%} & -54.83\%                  & -100.00\%           & 2.61                         \\
\textbf{\includegraphics[height=10.pt]{icons/boom.png}  0.25\%} & -55.07\%                  & -100.00\%           & 2.60                         \\
\toprule
\multicolumn{4}{c}{\textbf{\textit{Llama 3.2 3B Instruct}}}                                                        \\
\midrule
\textbf{Original}       & 58.93                     & 94.51               & -                            \\
\textbf{\includegraphics[height=10.pt]{icons/boom.png}  0.0025\%} & -37.73\%                  & -83.87\%            & 3.41                         \\
\textbf{\includegraphics[height=10.pt]{icons/boom.png}  0.01\%} & -42.91\%                  & -92.90\%            & 3.34                         \\
\textbf{\includegraphics[height=10.pt]{icons/boom.png}  0.09\%} & -48.64\%                  & -100.00\%           & 3.19                         \\
\textbf{\includegraphics[height=10.pt]{icons/boom.png}  0.25\%} & -49.14\%                  & -100.00\%           & 3.15 \\
\bottomrule
\end{tabular}
}
\caption{Comparison of LLM performance after \textit{Coding Spot} parameter deactivation. GTC (\%) and Code Change (\%) show changes in accuracy for general and coding tasks. \( M_s \) measures task-specific impact, and \includegraphics[height=8.pt]{icons/boom.png} marks models with deactivated parameters.}
\label{tab:results2}
\end{table}

\subsection{Results}
\subsubsection{Main Results}
The results clearly demonstrate the critical role of the \textit{Coding Spot} in both task-specific and general task performance. As shown in Table \ref{tab:results1}, deactivating even a small percentage of the most crucial parameters caused a significant decline in performance across the benchmarks.

For code-related tasks, such as HumanEval, Llama 3.1 8B Instruct, which initially achieved a score of 97.56, saw its performance plummet to zero when 0.09\% or 0.25\% of the \textit{Coding Spot} parameters were deactivated. This sharp decline highlights the essential role of these parameters in maintaining coding proficiency and suggests that the \textit{Coding Spot} encapsulates critical knowledge for code generation, generalizable across different programming languages.

For general tasks, such as GSM8K (mathematical reasoning), performance similarly exhibited notable declines when \textit{Coding Spot} parameters were deactivated. This indicates that the parameters critical for coding tasks also contribute to broader cognitive functions, underscoring the polysemantic nature of the \textit{Coding Spot}. However, tasks such as HellaSwag (commonsense reasoning) were less affected, indicating that distinct neural components may govern different general tasks, reflecting a modular structure within LLMs.

\subsubsection{Impact of \textit{Coding Spot} on Performance}

Our further exploration provides insights into the intricate dynamics between task-specific and general performance upon \textit{Coding Spot} parameter deactivation. The monosemanticity score \( M_s \), crafted to evaluate the changes in output across specialized and general tasks, is sensitive to the extent of parameter removal:

\begin{equation}
M_s = \frac{\Delta \text{Coding Task Performance}}{1 + \Delta \text{General Task Performance}}
\end{equation}

Table \ref{tab:results2} illustrates that Llama 3.1 and 3.2 models attained their highest monosemanticity scores with a minimal deactivation (0.0025\%), signifying that even a small fraction of parameter alteration can disproportionately influence coding tasks while leaving general capabilities relatively unscathed. This suggests a high density of critically functional parameters in these models, reflecting precise and efficient organization of the \textit{Coding Spot}. Conversely, for CodeLlama, more extensive deactivation (0.25\%) achieved peak scores, which may be attributed to its robust coding specialization derived from extensive training on diverse code repositories. This hints at a broader parametric allocation for coding tasks, confirming the presence of a more extensive \textit{Coding Spot} compared to instruction-tuned models like Llama 3.1 8B.

Our findings reveal that \textit{Coding Spot} parameters play crucial roles in tasks requiring logical and numerical reasoning. This is evident from the performance drop in the GSM8K benchmark, where mathematical reasoning—a domain overlapping with coding capabilities—is assessed. While commonsense reasoning tasks like HellaSwag remained stable, the decline in complex, math-related tasks highlights the essential nature of \textit{Coding Spot} parameters. These parameters not only enhance coding abilities but also support high-level problem-solving tasks, emphasizing their versatile roles within LLMs across broader cognitive functions.

\section{Conclusion}
We introduced the \textit{Coding Spot}, a specialized parameter subset in LLMs essential for both code generation and general tasks. Our experiments demonstrate that deactivating even a small percentage of these parameters leads to significant performance declines, confirming their critical role. These findings suggest that the \textit{Coding Spot} supports multiple domains, offering valuable insights for future work on optimizing LLM architectures to enhance both task-specific and general capabilities.

\section*{Limitations}

While our study provides significant insights into the structure and function of Large Language Models (LLMs), certain limitations must be acknowledged transparently. One of the primary limitations is the empirical selection of the threshold percentage \(k\%\) for identifying the \textit{Coding Spot}. Although empirical processes are commonly utilized to approximate optimal configurations when theoretical guidance is lacking, we recognize this approach may not guarantee absolute optimality across all LLM architectures. Future research could benefit from developing more robust, mathematically grounded methods for threshold determination.

Additionally, our methodology involves nullifying the \textit{Coding Spot} by setting the parameters to zero. Although this strategy effectively isolates the impact of these parameters, it might raise questions about interpretability, given the inherently positive and negative distribution of parameter weights. Setting them to zero provides a clear baseline for assessing their absence. However, we acknowledge this raises potential questions about non-zero alternatives, which might lead to different and perhaps unexpected model dynamics. While this alternative strategy has dividends in preserving network activity, introducing non-zero values could result in uncontrolled variance and divergent behaviors, rendering the results less interpretable. Thus, while this is an interesting avenue for future exploration, particularly for understanding robustness and sensitivity, the decision for zeroing parameters remains justified given current interpretability and controllability needs.

Our study was conducted exclusively using Llama models. This decision was intentional, designed to facilitate direct intra-architecture comparisons. A consistent model framework minimizes extraneous variability, allowing for a more focused analysis of parameter significance across different conditions. While this approach inherently limits our findings' generalizability to non-Llama architectures, it ensures robust internal comparison and serves as a strong foundation for future studies that can expand to more diverse LLM families.

While these constraints may be seen as potential shortcomings, we view them as informed choices given the current study's scope and analytical goals. They provide a baseline upon which future refinements and broader model inclusivity can be built.

\section*{Ethics Statement}
This research adheres to ethical guidelines in both the design and execution of experiments. The LLMs evaluated in this study were trained using publicly available data, and no private or sensitive information was involved. However, we acknowledge that LLMs, including those optimized for code generation, can raise concerns regarding fairness, bias, and security. It is important that future applications of this research take into account potential risks related to the misuse of automated code generation tools, especially in safety-critical contexts. We encourage further research on addressing these ethical concerns and ensuring the responsible deployment of LLM technologies.

\bibliography{custom}

\newpage

\appendix

\section{Related Work}
\label{appen:related_work}
Research on the coding capabilities of LLMs has gained significant attention, with models such as Llama 3 \cite{dubey2024llama}, GPT-4o \cite{achiam2023gpt}, and Claude 3.5 Sonnet \cite{Anthropic_2024} demonstrating strong performance in generating code across various programming languages. These models have successfully automated programming tasks by producing syntactically correct and logically coherent code. However, the underlying mechanisms that enable such capabilities remain unclear, particularly regarding how coding knowledge is distributed and organized within the model’s parameters. Understanding this parametric specialization is key to improving LLMs' ability to handle tasks such as code generation and comprehension.

LLM interpretability is a critical area of research aimed at uncovering how models process and store knowledge. Traditional techniques, such as saliency maps and gradient-based approaches \cite{simonyan2014very, sundararajan2017axiomatic}, have provided insights into neural network behavior by identifying feature importance. However, the scale and complexity of LLMs pose unique challenges in pinpointing the specific parameter subsets responsible for coding proficiency. This limitation has motivated the development of targeted methods for understanding the internal mechanisms of LLMs, especially in specialized tasks like code generation.

Our work draws inspiration from theories of modularity in both biological systems and artificial models. In cognitive science, the concept of modularity suggests that certain brain regions specialize in distinct functions, such as Broca’s and Wernicke’s areas for language processing \cite{fodor1983modularity}. Similarly, modularity has been proposed in neural networks, where specific neurons or regions are believed to specialize in particular tasks \cite{andreas2016neural}. Building on this analogy, we introduce the \textit{Coding Spot}, a specialized parametric region within LLMs dedicated to handling programming tasks. By identifying such parametric specialization, we aim to provide deeper insights into the internal architecture of LLMs and their proficiency in code-related tasks.

Recent advancements in parameter efficiency techniques, such as adapters \cite{houlsby2019parameter} and parameter-efficient fine-tuning methods \cite{zaken2021bitfit}, highlight the growing interest in improving model adaptability without retraining the entire network. These methods emphasize the importance of selectively fine-tuning critical parameter subsets for task-specific performance. Our work contributes to this field by offering a structured approach to identifying and optimizing task-critical parameters, specifically for coding tasks within LLMs.

Additionally, the concepts of monosemantic and polysemantic neurons \cite{elhage2022toy} have shaped our understanding of parameter specialization. Monosemantic neurons, which respond to specific stimuli, provide a framework for investigating parameter contributions within LLMs. In the context of code generation, identifying these specialized neurons or parameters is crucial for understanding task-specific organization. Our proposed \textit{Coding Spot} builds on this concept, positing that certain parameter subsets are responsible for coding tasks while preserving general non-coding functionalities.

In summary, our research builds on existing work in LLM interpretability, modularity, and parameter efficiency. By introducing the \textit{Coding Spot}, we bridge theoretical insights from cognitive science with practical advances in artificial model specialization, offering a novel framework for understanding and optimizing the coding capabilities of LLMs.

\section{Experimental Setup}
\label{appen:experimental_setup}

\subsection*{Datasets}
To evaluate the performance of our methodology, we use the \texttt{nampdn-ai/tiny-codes} \cite{nam_pham_2023} dataset, which includes Bash, C\#, C++, Go, Java, JavaScript, Julia, Ruby, Rust, and TypeScript. This dataset offers a diverse environment for fine-tuning LLMs on language-specific benchmarks, allowing a comprehensive evaluation of coding capabilities. The dataset includes real-world code samples for assessing both code generation and comprehension tasks within each language.

During preprocessing, non-code content was filtered out, and language-specific tokenization was applied to retain the syntactic and semantic integrity of each programming language, ensuring an accurate evaluation of the models.

It is critical to remove non-coding content, as our focus is on discerning the specific parameters that contribute to coding prowess. Inclusion of text related to instruction-following abilities or general reasoning could inadvertently skew the importance scores, misleadingly suggesting the relevance of parameters that are not genuinely integral to coding functions. This rigorous filtering guarantees that our analysis truly isolates the \textit{Coding Spot} responsible for coding, avoiding confounding factors associated with unrelated linguistic tasks.

\subsection*{Languages and Benchmarks}

In our study, we selected a diverse set of programming languages—Bash, C\#, C++, Go, Java, JavaScript, Julia, Ruby, Rust, and TypeScript. This selection covers a broad spectrum of programming paradigms, providing a comprehensive evaluation of the models' abilities across various coding styles and syntax.

We employed the HumanEval \cite{chen2021evaluating} benchmark, which is widely recognized for testing the Large Language Models' capability to generate code that is both syntactically correct and functionally coherent. This benchmark offers a structured framework for evaluating the proficiency of models in managing diverse coding tasks.

In addition to the coding benchmarks, our study included several other datasets aimed at assessing models on a variety of reasoning and task-specific abilities. The GSM8K \cite{cobbe2021training} benchmark was used to evaluate mathematical reasoning capabilities, challenging models to solve arithmetic and logic problems effectively. HellaSwag \cite{zellers2019hellaswag}, a benchmark designed for commonsense reasoning, offered insights into the LLMs' ability to navigate narrative completion tasks with contextual understanding.

For multi-task evaluation, we integrated MMLU \cite{hendrycks2020measuring}, which tests models on a wide range of academic and professional subjects, offering a holistic view of their reasoning and knowledge synthesis across disciplines. TruthfulQA \cite{lin2021truthfulqa} was employed to assess the truthfulness of model responses, pushing LLMs to adhere closely to factuality and truth in their outputs. Lastly, WinoGrande \cite{sakaguchi2021winogrande} tested coreference reasoning, emphasizing the ability to resolve pronouns based on contextual cues, thereby evaluating linguistic and contextual coherence.

These non-coding benchmarks collectively provide a rigorous and diverse assessment framework, ensuring a comprehensive evaluation of the LLMs beyond code generation, highlighting their versatility and adaptability across different reasoning scenarios and tasks.

\subsection*{Models}

Our experiments are conducted using three state-of-the-art models: CodeLlama 7B Instruct \cite{roziere2023code}, Llama 3.1 8B Instruct \cite{dubey2024llama}, and Llama 3.2 3B Instruct \cite{dubey2024llama}. These models were selected for their distinct capabilities and architectural strengths, providing a diverse set of test subjects for our study on parametric specialization.

The CodeLlama 7B Instruct model is particularly noteworthy for its expertise in code-related tasks. It is pre-trained on a vast corpus of coding-specific data, enabling it to navigate complex coding challenges with precision and efficiency. Its architecture is fine-tuned for tasks that require generating syntactically and semantically accurate code, making it an ideal candidate for testing the \textit{Coding Spot} hypothesis in environments demanding high-level coding proficiency.

On the other hand, Llama 3.1 8B Instruct and Llama 3.2 3B Instruct offer robust performance across a broader range of instructional content beyond pure coding, including natural language understanding and generalized reasoning. These models provide insight into how the \textit{Coding Spot} can extend its specialized responsibilities across varied cognitive demands. Their sizeable parameter sets and nuanced training regimes allow them to adapt well to fine-tuning on language-specific datasets, providing a comprehensive view of parameter specialization and monosemantic behavior across different scales and instruction types.

\end{document}